\documentclass{ifacconf}

\usepackage{url}
\usepackage{textcomp, gensymb} % Extra symbols
\usepackage{placeins} % FloatBarrier definition
\usepackage{stfloats}% vrorce 32-44 zarovnaný nahoru
\usepackage{cool}
\usepackage{natbib}

\usepackage[utf8]{inputenc}
\usepackage{graphicx}

\usepackage{amsmath}
\usepackage{amssymb}
\usepackage{mathrsfs}
\usepackage{bm}
\usepackage{siunitx}
\usepackage{cancel}
\usepackage[scr=boondoxo,scrscaled=1.05]{mathalfa} % Math alphabets, such as \mathbb

\usepackage{algorithm}% http://ctan.org/pkg/algorithms
\usepackage{algpseudocode}% http://ctan.org/pkg/algorithmicx

\usepackage{multirow}
\usepackage{multicol}
\usepackage{tabularx}
\usepackage{booktabs} 
\usepackage{hhline}

\usepackage{commands}
\usepackage{definice}

\newcommand{\tightoverset}[2]{%
	\mathop{#2}\limits^{\vbox to -.5ex{\kern-0.75ex\hbox{$#1$}\vss}}}

\makeatletter
\def\endfigure{\end@float}
\def\endtable{\end@float}
\makeatother

\let\ifacconfcaptionwidth\captionwidth
\usepackage[caption=false]{subfig}
\let\captionwidth\ifacconfcaptionwidth
\begin{document}
	\begin{frontmatter}
		
		\title{\hspace{-1mm}Online Learning and Control for Data-Augmented Quadrotor Model\hspace{-0mm}}
		
		\author[First]{Matěj Šmíd}
		\author[Second]{Jind\v{r}ich Duník}

		\address[First]{Institute of Computer Science, Czech Academy of Sciences, Prague, Czech Republic\\ Dept. of Cybernetics, University of West Bohemia, Pilsen, Czech Republic 
		(e-mail: smid@cs.cas.cz)}
		\address[Second]{Dept. of Cybernetics, University of West Bohemia, Pilsen, Czech Republic (e-mails: \{matoujak,dunikj\}@kky.zcu.cz).}
		
		\begin{abstract} % Abstract of not more than 250 words.
			The ability to adapt to changing conditions is a key feature of a successful auto\-nomous system. 
			In this work, we use the Recursive Gaussian Processes (RGP) for identification of the quadrotor air drag model \emph{online}, without the need to precollect training data. 
			The identified drag model then augments a physics-based model of the quadrotor dynamics, which allows more accurate quadrotor state prediction with increased ability to adapt to changing conditions. 
			This \emph{data-augmented physics-based} model is utilized for precise quadrotor trajectory tracking using the suitably modified Model Predictive Control (MPC) algorithm. 
			The proposed modelling and control approach is evaluated using the Gazebo simulator and it is shown that the proposed approach tracks a desired trajectory with a higher accuracy compared to the MPC with  the non-augmented (purely physics-based) model. 
			\\\\
			{{\emph{Source code}: }\url{https://github.com/smidmatej/mpc_quad_ros}}
            
			\vspace*{-3mm}
		\end{abstract}
		
		
	\end{frontmatter}
	%===============================================================================
	
	\showboxdepth=\maxdimen
\showboxbreadth=\maxdimen
	\section{Introduction}
	Given a model of a dynamical system, we seek to find discrepancies between the model and the actual system. This is a common problem in control theory, where the dynamics model is often a simplification of the actual system. To this end, machine learning (ML) models are a popular choice to model such discrepancies and for subsequent augmentation of pure physics-based dynamics models \citep{carron2019data,kabzan2019learning,hewing2019cautious,torrente2021data}. 
	\let\thefootnote\relax\footnotetext{© 2024 the authors. This work has been accepted to IFAC for publication under a Creative Commons Licence CC-BY-NC-ND.}
    
	We present an approach to control a quadrotor using a first principles model augmented with an Recursive Gaussian Process (RGP) \citep{huber2014recursive,wahlstrom2015extended} model that serves to describe
	the quadrotor's air drag characteristics. 
	The augmented model is used in a Model Predictive Control (MPC) framework to allow for precise trajectory tracking accounting for the quadrotor's air drag without assuming a priori knowledge of the air drag form or coefficients. 
	ML augmented models typically require a training dataset to be collected in a separate run which is then used to fit the ML model \citep{torrente2021data, GP3DdroneMehndiratta, Jin_Wu_Liu_Zhang_Yu_2021}. 
	In contrast, we fit the RGP model online at each time step in a recursive fashion. This allows us to adapt to changing environmental conditions \emph{without} the need to collect a new training dataset. Recently, a similar approach was developed, making use of state-space implementation of GP to achieve real time GP control \citep{Schmid_Gruner_Abbas_Rostalski_2022}.

	Combining a GP model with MPC control is not a novel idea, it has been used in the past for control of generic non-linear models \citep{Cao_Lai_Alam_2017}, quadrotors \citep{cao2017gaussian}, and UAV platforms \citep{Hemakumara_Sukkarieh_2011}, with the main draw being the reduced labor required to construct GP models compared to first principles Newtonian models.

	The rest of the paper is organized as follows.
	Section \ref{sec:problem_formulation} describes the first principles quadrotor model and assumptions about its drag characteristics.
	In Section \ref{sec:motivation} we give a high level overview of the proposed approach, seen through the lens of \emph{data-augmented-physics-based} modeling.
	Section \ref{sec:data-aug-mpc} describes in detail the proposed method for observing the drag, using those observations to train the RGP regression, augmenting the first principles model with the RGP, and using this augmented model in the MPC framework.
	Section \ref{sec:simulation-experiments} describes the simulation environment used for evaluation, the specificities of the conjunction of the MPC with the RGP, and the results.
	Finally, Section \ref{sec:conclusions} summarises both the limitations and the possible future research paths of the proposed approach. 
	
	%%%%%%%%%%%%%%%%%%%%%%%%%%%%%%%%%%%%%%%%%%%%%%%%%%%%%%%%%%%%%%%%%%%%%%%%%%%%%%%%
	\begin{figure}[t!]
		\centering
		\includegraphics[width=\linewidth]{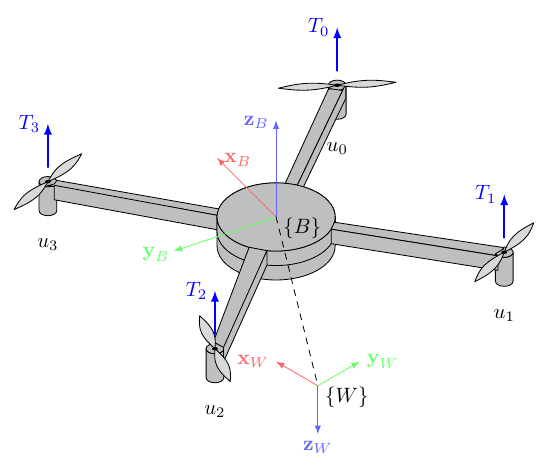}
		\caption{Quadrotor schematic. Thrust $T_i$ is generated by individual rotors given inputs $u_i$ The body frame $\left\{ x_B, y_B, z_B \right\}$ is attached to the center of mass of the quadrotor. The world frame $\left\{x_W, y_W, z_W \right\}$ is attached to the inertial frame.}
		\label{fig:drone}
	\end{figure}

	\section{PHYSICS-BASED QUADROTOR MODELLING}
	\label{sec:problem_formulation}
	
	\subsection{Notation}
	To describe the quadrotor, we need to consider 2 reference frames. The world reference frame $W$ is defined by the orthonormal basis $\left\{ \bm{x}_W, \bm{y}_W, \bm{z}_W \right\}$ and the body reference frame $\Bodyframe/$ by the basis vectors $\left\{\bm{x}_B, \bm{y}_B, \bm{z}_B \right\}$. 
	The origin of $\Bodyframe/$ is colocated in the center of mass of the quadrotor. 
	We denote the vector from coordinate system $\Worldframe/$ to coordinate system $\Bodyframe/$, as seen from $\Bodyframe/$ as $\bm{v}^B$, and $\bm{v}^W$ as $\Bodyframe/ \rightarrow \Worldframe/$ as seen from $\Worldframe/$.
	The quaternion--vector product is denoted by $\odot$, representing the rotation of $\bm{v}$ by $\bm{q}$. It is defined as $\bm{q} \odot \bm{v} = \bm{q} \bm{v} \bm{q}^*$, where $\bm{q}^*$ is the quaternion conjugate of $\bm{q}$. The cross--product of vectors $\bfa, \bfb$ is denoted as $\bfa\times\bfb$.
	
	\subsection{Quadrotor Model}

	Following \citep{torrente2021data}, we assume a quadrotor can be described as a rigid body with 6-DoF in free space. The state of the quadrotor $\bm{x}$ is chosen as the quadrotor current position $\bm{p}^W = \left[x,y,z\right]$, rotation quaternion
	%\footnote{To describe rotation we use a quaternion  describing the rotation of frame $\Bodyframe/$ relative to frame $\Worldframe/$. Rotation of a object in $\mathbb{R}^3$ is described fully by 3 angles (such as using Euler angle representation), but the 4-dimensional quaternion with length 1 allows us to avoid the gimbal lock problem present in other representations.}
	$\bm{q}^{W} = \left(q_w, q_x, q_y, q_z \right)$, velocity $\bm{v}^W = \left[v_x,v_y,v_z \right]$ and angular rate $\bm{\omega}^W = \left[\dot{\varphi}, \dot{\theta}, \dot{\psi} \right]$.
	
	The quadrotor's rotors are collocated on the $xy$-plane of the $\{B\}$ frame, which is centered in the center of mass of the quadrotor. The rotors are numbered $0$ to $3$ in a clockwise direction when seen from the positive $\bm{z}_B$ direction. Furthermore, the $\bm{x}_B$ axis is pointing forward, in between the zeroth and the third rotor as seen in \figref{fig:drone}.
	
	The model is controlled by an input vector $\bm{u} = \left[u_0, u_1, u_2, u_3 \right]$, where $u_i \in [0,1]$ for $i = 0,1,2,3$. The input $\bm{u}$ corresponds to the activation of each individual rotor. Each rotor provides a thrust $T_i = T_{\text{max}} \cdot u_i$. The collective thrust vector $\bm{T}^B$ and the torque vector $\bm{\tau}^B$, both expressed in the body frame $\Bodyframe/$ and constituted by the individual thrusts $T_i$, are given by 
	\begin{align}
		\bm{T}^B = \begin{bmatrix} 0 \\ 0 \\ \sum_{i} T_i \end{bmatrix}
		\; \text{and} \; 
		\bm{\tau}^B = \begin{bmatrix} 
			d_y (-T_0 - T_1 + T_2 + T_3) \\
			d_x (-T_0 + T_1 + T_2 - T_3) \\ 
			c_{\tau} (-T_0 + T_1 - T_2 + T_3) \\ 
		\end{bmatrix},
		\label{eq:thrustandtorque}
	\end{align}
	where $d_x, d_y$ are the rotor displacements and $c_{\tau}$ is the rotor drag constant.

	We model the nominal quadrotor dynamics using the following a 13-dimensional state vector $\bm{x}$ with corresponding ordinary differential equations (ODEs) \cite{Furrer2016}
	\begin{align}
		\dot{\bm{x}} = 
		\fphys/ = 
		\begin{bmatrix} 
			\dot{\bm{p}}^W \\
			\dot{\bm{q}}^W \\
			\dot{\bm{v}}^W \\
			\dot{\bm{\omega}}^W 
		\end{bmatrix}
		=
		\left[\begin{smallmatrix} 
			\bm{v}^W  \\
			\bm{q}^W \cdot \left[\begin{smallmatrix} 0 \\ \bm{\omega}^B /2 \end{smallmatrix}\right] \\
			\bm{q}^W \odot \left( \frac{1}{m} \bm{T}^B \right) + \bm{g}^W \\
			\bm{J}^{-1} (\bm{\tau}^B - \bm{\omega}^B \times \bm{J} \bm{\omega}^B) 
		\end{smallmatrix}\right].
		\label{eq:quadrotor_dynamics}
	\end{align}

	The parameters of the model are the mass $m$, moment of inertia $\bm{J} \in \mathrm{R}^{3\times3}$ and the gravitational acceleration vector $\bm{g}_{W} = [0, 0, -9.81]^T ms^{-2}$.

	%We model the nominal quadrotor dynamics using the following ordinary differential equations (ODEs) \cite{Furrer2016}
	%\begin{align}
	%\begin{split}
	%    \dot{\bm{p}}^W &= \bm{v}^W \\
	%    \dot{\bm{q}}^W &= \bm{q}^W \cdot \begin{bmatrix} 0 \\ \bm{\omega}^B /2\end{bmatrix} \\
	%    \dot{\bm{v}}^W &= \bm{q}^W \odot \left( \frac{1}{m} \bm{T}^B \right) + \bm{g}^W \\
	%    \dot{\bm{\omega}}^W &= \bm{J}^{-1} (\bm{\tau}^B - \bm{\omega}^B \times \bm{J} \cdot \bm{\omega}^B).
	%    \label{eq:quadrotor_odes}
	%\end{split}
	%\end{align}
	%
	%
	%The parameters of the model are the mass $m$, moment of inertia $\bm{J} \in \mathrm{R}^{3\times3}$ and the gravitational acceleration vector $\bm{g}_{W} = [0, 0, -9.81]^T ms^{-2}$.
	%
	%
	%Together, these ODEs describe the \emph{physical} model of the quadrotor dynamics using a 13-dimensional state vector $\bm{x}$ as 
	%\begin{align}
	%    \dot{\bm{x}} = \fphys/ = \left[\begin{smallmatrix} \bm{v}_{WB} \\ \bm{q}_{WB} \cdot \left[\begin{smallmatrix} 0 \\ \bm{\omega}_B /2\end{smallmatrix}\right] \\  \bm{q}_{WB} \odot \left( \frac{1}{m} \bm{T}_B \right) + \bm{g}_W \\ \bm{J}^{-1} (\bm{\tau}_B - \bm{\omega}_B \times \bm{J} \bm{\omega}_B) \end{smallmatrix}\right].
	%    \label{eq:quadrotor_dynamics}
	%\end{align}

	\subsection{Drag Modelling}
	Drag is a force acting on a body counter to its motion in a fluid. One source of drag is solid's body drag, which is proportional to the velocity of the body and the fluid density. The body drag force acting on the quadrotor body is described as 
	\begin{align}
		\bm{F}^B_{bd} = -\frac{1}{2} \rho \vert \bm{v}^B \vert ^2 \coefdrag/ A  \bm{v}^B,
		\label{eq:drag_body}
	\end{align}
	where $\rho$ is the fluid density, $\coefdrag/$ is the drag coefficient and $A$ is the cross-sectional area of the body. The drag force is applied to the body in the opposite direction of the velocity vector $\bm{v}^B$.
	
	Another source of drag affecting the quadrotor is the rotor drag affecting each rotor which can be modelled using 
	\begin{align}
		\bm{F}^B_{rd} = - \omega C_{rD} { \left(\bm{v}^B \right)}^{\perp},
		\label{eq:drag_rotor}
	\end{align}
	where $\omega$ is the angular velocity of the rotor, $C_{rD}$ is the drag coefficient of the rotor and $\left(\bm{v}^B \right)^\perp$ is  the orthogonal projection of the quadrotor $\Bodyframe/$ velocity  onto the $xy$-plane.
	
	Both drag equations, i.e., \eqref{eq:drag_body} and \eqref{eq:drag_rotor} describe a force and therefore an acceleration affecting a body. In the case of our quadrotor model, this directly affects the change in velocity $\bm{v}^W$ in \eqref{eq:quadrotor_dynamics}. Since we do not have access to direct measurements of the experienced drag force, we will use the measured acceleration as a surrogate and work in the acceleration domain for the rest of the paper.
	\section{Motivation, Related Work, Goal of the Paper}
	\label{sec:motivation}
	The drag model parameters, such as the coefficients, fluid density, or the quadrotor-related areas, and their spatial and time variability, are, unfortunately, known with a very \emph{limited} accuracy. To improve the accuracy, the model of the total drag acceleration or its parameters should be identified from measured data, while the physical model \eqref{eq:quadrotor_dynamics} is respected. 
	This, recently developed concept, is referred to as the \emph{data-augmented physics-based} (DAPB) modelling \citep{Imbiriba}.

%@INPROCEEDINGS{Imbiriba,
%  author={Imbiriba, Tales and Demirkaya, Ahmet and Duník, Jindřich and Straka, Ondřej and Erdoğmuş, Deniz and Closas, Pau},
%  booktitle={2022 25th International Conference on Information Fusion (FUSION)}, 
%  title={{Hybrid Neural Network Augmented Physics-based Models for Nonlinear Filtering}}, 
%  year={2022},
%  volume={},
%  number={},
%  pages={1-6},
%  doi={10.23919/FUSION49751.2022.9841291}
%}

	In \citep{torrente2021data, GP3DdroneMehndiratta}, the total acceleration error made up by both drag forces\footnote{Since the relationship between acceleration and force is given by Newton's second law, it is not possible to directly disentangle the effects of the body and rotor drag from each other.} \eqref{eq:drag_body}, \eqref{eq:drag_rotor}, was modelled using an offline trained Gaussian process (GP) which was then used as an extension to the physical model for a trajectory control. The approach leads to significant improvement of the resulting position accuracy assuming \emph{no} time or spatial variability of the drag-related acceleration, a condition which might not be always fulfilled.
	
	The goal of this paper is to further extend and simplify the applicability of the DAPB quadrotor modelling by online identification of the drag-related acceleration. In particular, we
	\begin{itemize}
		\item Use \emph{recursive} GP for online drag modelling,
		\item Design a MPC algorithm capable of working with inherently time-varying DAPB models.
	\end{itemize}
	We illustrate the viability of the proposed method using a realistic scenarios generated Gazebo simulator \citep{koenig2004design} and publicly available source code.

	\section{DATA-AUGMENTED QUADROTOR MODEL AND PREDICTIVE CONTROL}
	\label{sec:data-aug-mpc}
	
	\subsection{Drag Estimation}
	Inspired by previous work \citep{torrente2021data}, we model the total drag force given by \eqref{eq:drag_body}, \eqref{eq:drag_rotor} by estimating the drag acceleration $\tilde{\bm{a}}^B$ acting on the quadrotor in the body frame $\Bodyframe/$ at time $t_k$ as
	\begin{align}
		\tildea/ = \frac{\bm{v}^B_{k+1} - \hat{\bm{v}}^B_{k+1}}{\Delta t_k},
		\label{eq:drag_body_acc}
	\end{align}
	where $\hat{\bm{v}}^B_{k+1}$ is the velocity predicted using the nominal model $\fphys/$, $\bm{v}^B_{k}$ is the measured velocity, $\Delta t_k = t_k - t_{k-1}$, and using the notational shorthand $\bm{v}^B_{k}=\bm{v}^B(t_k)$. The estimated drag acceleration $\tildea/$, together with the current velocity $\bm{v}^B_k$ then form an observation pair 
	\begin{align}
		\ok{k} = \left\{\bm{v}^B_{k}, \tildea/ \right\}, \label{eq:pair_o}
	\end{align}
	that is used in the regression procedure described below.
	
	Using $\tildea/$ as a surrogate for model discrepancy is not a obvious choice, a different surrogate might be more appropriate, depending on the model in question and the assumed unmodelled disturbances. 
	
	\subsection{Drag Modelling by Recursive Gaussian Process}
	
	Gaussian process regression is an alternative data-driven technique
	based on Bayesian theory for fitting a probability distribution of functions to available data, providing immidiate access to uncertainity information when running inference using the regressed model \citep{rasmussen2005gaussian}.

	The standard  ''offline'' GP require access to $\nobs/$ observations $\ok{k} = \left\{\bm{v}^B_{k}, \tilde{\bm{a}}^B_{k} \right\}$ forming the dataset $\dataset/ = \left\{ \ok{k} \; \vert \; k = 0,\ldots, \nobs/ \right\}$ gathered from the quadrotor during flight. 
	To reduce the complexity of the GP we split the observations into its $d \in \{x,y,z\}$ components, use a separate GP for each component and then aggregate their outputs into an ensemble.
	The observations $\ok{k}$ forming the dataset $\dataset/$ is generated from the noisy process 
	\begin{align}
		\tilde{a}^B_{d,k} = \gd/(v^B_{d,k}) + \epsilon_{d,k},
		\label{eq:gp_obs}
	\end{align}
	where $\gd/$ is the true, unmodelled drag acceleration for dimension $d$ and $\epsilon_{d,k} \sim \mathcal{N}(0, \sigma_d^2)$ is the Gaussian noise in the measurement. The GP is used to infer the latent function $\gd/$ from  $\dataset/$. Learning in this context involves a process of finding the hyperparameters $\eta$ minimizing the RMSE of fit for the available dataset $\dataset/$.

	\subsection{Recursive Gaussian Process Regression}
	
	Recursive Gaussian process regression \citep{huber2014recursive} is a method for \emph{online} learning of Gaussian processes, allowing for the expansion of the set of available observations as new observations $\ok{k}$ become available. 
	
	The RGP for each dimension $d \in \{x,y,z\}$ is initialized with a set of basis vectors located at $\vbasis/ \triangleq [ v^+_{\ddim/,1}, v^+_{\ddim/,2}, \ldots, v^+_{\ddim/,m}]$ with the random variable $\abasis/ \triangleq \left[ \tilde{a}^+_{\ddim/,1}, \tilde{a}^+_{\ddim/,2}, \ldots, \tilde{a}^+_{\ddim/,m} \right]$, corresponding to output of $\gd/$ given a velocity $v^+_{\ddim/,k}$.  
	These basis vectors can be thought of as an active set allowing a sparse GP representation. This alleviates the computational requirements immensely, since $m \ll n_{\text{obs}}$.
	%  = \left[ f_{\text{RGP},\ddim/}(v^+_{\ddim/,1}), f_{\text{RGP},\ddim/}(v^+_{\ddim/,2}), \ldots, f_{\text{RGP},\ddim/}(v^+_{\ddim/,m}) \right]
	The initial distribution of the RGP is Gaussian with its probability density function (PDF) $\xi(x \vert \mu, \sigma)$ being 
	\begin{align}
		p_0(\abasis/) = \xi(\abasis/ \vert \mubasiskd{0}{\ddim/}, \Cbasiskd{0}{\ddim/}),
		\label{eq:initial_rgp}
	\end{align}
	where $\mubasiskd{0}{\ddim/} \triangleq \bm{0}$ is the mean at $\vbasis/$ and $\Cbasiskd{0}{\ddim/} \triangleq \kernel/(\vbasis/, \vbasis/)$\footnote{$\kernel/(x,x') = \sigma_f^2 \exp \left(-\frac{1}{2} \frac{(x-x')^2}{l} \right) + \sigma^2_n$ is the squared exponential kernel. When applied to vector arguments, the kernel $\kernel/$ is applied element-wise to each pair of elements in the arguments, yielding a matrix. The hyperparameters $\eta = \gphyperparams/$ are the only hyperparameters of the GP/RGP regression.} is the covariance at $\vbasis/$.

	The goal is then to use the new observations $\ok{k}$ as they become available to calculate the posterior distribution $p(\abasis/ \vert \okt{k})$, where $\okt{k} \triangleq \{\ok{1}, \ok{2}, \ldots, \ok{k}\}$, from the prior distribution
	\begin{align}
		p(\abasis/ \vert \okt{k-1}) = \xi(\abasis/ \vert \mubasiskd{k-1}{\ddim/}, \Cbasiskd{k-1}{\ddim/}).
		\label{eq:posterior_rgp}
	\end{align}
	At any point in time, we can run inference to find the distribution of $\aarb/ = \gd/(\varb/)$, i.e., of the function $\gd/$ at an arbitrary point $\varb/$, in our case corresponding to the drag acceleration given an instantaneous velocity, by evaluating the joint prior $p(\abasis/, \aarb/ \vert \okt{k})$ and marginalizing out the basis vectors $\abasis/$ such as
	\begin{gather}
		\begin{aligned}
			p(\abasis/, \aarb/ \vert \okt{k}) &= p(\aarb/ \vert \abasis/) p(\abasis/ \vert \okt{k})\\
			&= \xi(\aarb/ \vert \muarb/, \Carb/) \xi(\abasis/ \vert \mu^{+,k}_{\ddim/}, \bm{C}^{+,k}_{\ddim/}),
			\label{eq:joint_rgp}
		\end{aligned}
	\end{gather}
	where
	\begin{align}
		&\muarb/ \triangleq \bm{H} \cdot \mubasiskd{k}{\ddim/}, \\
		&\Carb/ \triangleq \kernel/(\varb/, \varb/) - \bm{H} \cdot \kernel/(\varb/, \vbasis/), \\
		&\bm{H} \triangleq \kernel/(\varb/, \vbasis/) \cdot \kernel/(\vbasis/, \vbasis/)^{-1}.
	\end{align}
	
	\subsection{Model Learning, Utilisation, and Properties}
	We use the RGP model to compensate the discrepancy between our nominal model prediction and the measured state of the quadrotor characterised by the drag-related acceleration \eqref{eq:drag_body_acc}.
	
	For each RGP in the three element ensemble, the basis vectors $\vbasis/$ are initialized by equidistantly sampling the velocity space on $\left[-v_{max}, v_{max} \right]$ using $m$ points, setting their corresponding estimates  of $\gd/$ to $0$ and their covariance to $\bm{C}^{+,0}_{d} = \kernel/(\vbasis/, \vbasis/)$.
	At each time step $k$, we use the current quadrotor state $\bm{x}_k$ with the RGP model $\frgp/$ to infer the acceleration $\tilde{\bm{a}}$ at the current velocity $\bm{v}_k$ using the mean of the marginalized joint distribution in \eqref{eq:joint_rgp} as
	\begin{align}
		\frgp/( \bm{x}_k ; \muarbbold/ )= 
		\begin{bmatrix} 
			\bm{0}^{7 \times 1}\\
			\tilde{a}_x \\
			\tilde{a}_y \\
			\tilde{a}_z \\
			\bm{0}^{3 \times 1}.
		\end{bmatrix}
		=
		\begin{bmatrix} 
			\bm{0}^{7 \times 1}\\
			\muarbd{x} \left(v_{x,k} ; \; \mubasiskd{k}{x} \right)\\ 
			\muarbd{y} \left(v_{y,k} ; \; \mubasiskd{k}{y} \right)\\ 
			\muarbd{z} \left(v_{z,k} ; \; \mubasiskd{k}{z}\right)\\
			\bm{0}^{3 \times 1}.
		\end{bmatrix},
		\label{eq:rgp_model}
	\end{align}
	where $\muarbbold/ \triangleq \left(\muarbd{x}, \muarbd{y}, \muarbd{z} \right)$.
	To perform the online update of the RGP distribution $p(\abasis/ \vert \okt{k})$ we keep the covariance matrices $\Cbasiskd{k-1}{\ddim/}$ and the mean $\mubasiskd{k-1}{\ddim/}$ of the prior distribution $p(\abasis/ \vert \okt{k-1})$. However, to run inference using the RGP, only the means $\mubasiskd{k}{\ddim/}$ are needed, while $\Cbasiskd{k-1}{\ddim/}$ is kept in memory only to perform updates.

	%
	%Given a velocity $\bm{v}_k^*$, the RGP model provides a prediction as a normal distribution ensemble of normal distributions.
	%We then use the mean of these distributions as the predicted acceleration $\bm{a}_{D,k}^*$. 
	%
	%The correction given by the prediction model is given by
	%\begin{align}
	%    \frgp/( \bm{x}_k^* ; \bm{\mu}^g_k) = \begin{bmatrix} 
		%        \bm{0}^{7 \times 1}\\
		%        \mu_{x,k}^* \left(v^*_{x,k} ; \bm{\mu}^g_{x,k} \right)\\ 
		%        \mu_{y,k}^* \left(v^*_{y,k} ; \bm{\mu}^g_{y,k} \right)\\ 
		%        \mu_{z,k}^* \left(v^*_{z,k} ; \bm{\mu}^g_{z,k} \right)\\
		%        \bm{0}^{3 \times 1}.
		%    \end{bmatrix}
	%\end{align}
	%Here, it is important to note the difference between $\bm{\mu}_{k}^*$ and $\bm{\mu}^g_{k}$. The value $\bm{\mu}_{k}^*$ is the mean of the inference distribution, this value is calculated during inference and it is not saved in the model. 
	%In contrast, $\bm{\mu}^g_{k}$ is the mean of the RGP at the basis vectors $\bm{X}$, which is updated during training and can be treated as a parameter of the RGP.
	%
	
	\subsection{Data-Augmented Physics-Based Model}
	To better model the quadrotor dynamics, the physics-based model $\fphys/$ from \eqref{eq:quadrotor_dynamics} and the online trained RGP model $\frgp/$ from \eqref{eq:rgp_model} can be combined into the single \emph{augmented} model
	\begin{align}
		\fpred/ \left( \bm{x}_k , \bm{u}_k \right) = \fphys/\left( \bm{x}_k , \bm{u}_k \right) + \frgp/ \left( \bm{x}_k; \rgpparam/ \right),
		\label{eq:pred_model}
	\end{align}
	which provides more accurate description of the underlying reality. It is important to note a limitation of this approach: the $\frgp/$ model is dependent only on the velocity of the quadrotor, however, the rotor drag force \eqref{eq:drag_rotor} is dependent on the angular velocity of the rotors ($\bm{u}_k$). This means that the RGP model is not able to capture the rotor drag force, only the body drag force.

	To use $\fpred/$ in discrete time algorithms, we use the Runge-Kutta 4th order (RK4) integration method
	\begin{align}
		\begin{split}
		\bm{x}_{k+1} =& \frk/(\bm{x}_k, \bm{u}_k, \Delta t) = \\
		&\bm{x}_k + \frac{\Delta t}{6} \left( \bm{k}_1 + 2\bm{k}_2 + 2\bm{k}_3 + \bm{k}_4 \right),
		\end{split}
		\label{eq:rk4}
	\end{align}
	where $\bm{k}_1, \bm{k}_2, \bm{k}_3, \bm{k}_4$ are the RK4 intermediate steps given calculated from $\fpred/$.

	\subsection{Model Predictive Control for Augmented Model} 
	Model predictive control \citep{magni2009nonlinear} is a control scheme used to stabilize a system subject to dynamic equations $\dot{\bm{x}} = \fpred/(\bm{x},
	\bm{u})$  with equality and inequality constraints $\bm{r}, \bm{h}$ given a reference trajectory $\xref/(t)$ by minimizing a cost function. 

	\if 0
	MPC solves the optimal control problem (OCP) defined as 
	\begin{align}
		\begin{split}
			&\underset{\bm{u}}{\text{min} } \int_t^{t+t_h} \mathcal{L}(\bm{x}, \xref/, \bm{u}, \uref/) d\tau \\
			\text{subject to } &\dot{\bm{x}} = \fpred/(\bm{x}, \bm{u}),\quad
			\bm{x}(t) = \bm{x}_0, \\
			&\bm{r}(\bm{x}, \bm{u}) = 0, \hspace*{12mm}
			\bm{h}(\bm{x}, \bm{u}) \leq 0,
			\label{eq:mpc}
		\end{split}
	\end{align}
	where $\fpred/$ denotes the known model of the system, $\bm{x}_{\text{ref}}$ is the tracked trajectory, $\bm{x}_0$ is the initial state, $t$ is the current time, $t_h$ is the horizon look-forward time,   
	$\bm{r}$ describes the equality constraints and $\bm{h}$ describes the inequality constraints.
	For simplicity, we chose to set the cost functional as $\mathcal{L}(\bm{x}, \xref/, \bm{u}, \uref/) = (\bm{x} - \xref/) \bm{Q} (\bm{x} - \xref/)^T$, where $\bm{Q} = \text{diag}([10, 10, 10, 0.1, 0.1, 0.1, 0.05, 0.05, 0.05, 0.05, 0.05, 0.05])$, i.e, a quadratic cost functional putting emphasis on the positional error. 
	\fi

	MPC solves the optimal control problem (OCP) defined as 
	\begin{align}
		\begin{split}
			&\underset{\bm{u}}{\text{min} } \sum_{k=0}^{N} \tilde{\bm{x}}_k^T \bm{Q} \tilde{\bm{x}}_k  + \bm{u}_k^T \bm{R} \bm{u}_k \\
			\text{subject to } &\bm{x}_{k+1} = \frk/(\bm{x}_k, \bm{u}_k),\quad
			\bm{x}_0 = \bm{x}_{\text{init}}, \\
			&\bm{r}(\bm{x}_k, \bm{u}_k) = 0, \hspace*{12mm}
			\bm{h}(\bm{x}_k, \bm{u}_k) \leq 0,
			\label{eq:mpc}
		\end{split}
	\end{align}
	where $\tilde{\bm{x}}_k = \bm{x}_k - \bm{x}_k^{\text{ref}}$, $\bm{x}_k^{\text{ref}}$ is the tracked trajectory, $\bm{x}_{\text{init}}$ is the initial state, $t$ is the current time, $t_h$ is the horizon look-forward time,   
	$\bm{r}$ describes the equality constraints and $\bm{h}$ describes the inequality constraints. 	We discretize the system $\fpred/$ into $N$ steps over time horizon $T$ of size $dt = T/N$. The state weight matrix is set to $\bm{Q} = \text{diag}([10, 10, 10, 1, 1, 1, 0.5, 0.5, 0.5, 0.5, 0.5, 0.5])$ and the control weight matrix is set to $\bm{R} = \text{diag}([0.1, 0.1, 0.1, 0.1])$.

	Given we want to update the control algorithm online with the new RGP model, it would be computationally inefficient to reconstruct the OCP at each control time step, therefore, we treat the vector $\rgpparam/$ as a parameter of the $\frgp/$ model. In this way, we change the parameters of the nonlinear optimization problem at each control time step $k$ as the RGP ensemble is updated. 
	The process of computing the control using MPC and updating its RGP model is shown in Algorithm \ref{alg:control}. We refer to this approach as \namergp/ and as \namegp/ when a pre-trained GP is used.

	\begin{algorithm}
		\caption{Recursive Learning and Control Algorithm Summary}
		\label{alg:control}
		\begin{algorithmic}[1]
			
			\State initialize $\frgp/$ with $\vbasis/$ and $\abasis/$
			\State initialize MPC with $\frk/$
			\State $\hat{\bm{x}}_0 \gets $ initial state
			\State $k \gets 0$
			\While{trajectory ready}
			\Procedure{\algprocname/}{} \label{alg:mpc}
			\State $\bm{x}_k \gets $ state measurement
			\State $\bm{u}_{k, 0:0+n_h} \gets $ MPC optimal control
			\State send control command $\bm{u}_{k,0}$ 
			\State $\hat{\bm{x}}_{k+1} \gets  \fphys/ \left(\bm{x}_k, \bm{u}_{k,0} \right)$ \eqref{eq:quadrotor_dynamics}
			\State $\tildea/ \gets $ \eqref{eq:drag_body_acc}
			\State $\ok{k} \gets \left\{\bm{v}^B_{k}, \tildea/ \right\}$
			\State $p(\abasis/ \vert \okt{k}) \gets $ update with $\ok{k}$ from prior
			\State $\rgpparam/ \gets$ marginalize $p(\abasis/, \tildea/ \vert \okt{k})$
			\State update $\frk/$ with $\rgpparam/$ inside MPC
			\State $k \gets k + 1$
			\EndProcedure
			\EndWhile
		\end{algorithmic}
	\end{algorithm}
	
	%%%%%%%%%%%%%%%%%%%%%%%%%%%%%%%%%%%%%%%%%%%%%%%%%%55
	\section{SIMULATION EXPERIMENTS}
	\label{sec:simulation-experiments}
	The proposed data-augmented modelling and and predictive control of quadrotor is numerically evaluated on two  trajectories using a realistic quadrotor model provided by the Gazebo simulator. 
	Details on MPC implementation, and the results follows. 
	
	%
	%\begin{figure}
	%    \begin{subfigure}{.23\textwidth}
		%      \centering
		%      \includegraphics[width=1\linewidth]{figures/trajectory_random.pdf}
		%      \caption{Trajectory created by interpolating between randomly generated waypoints with a polynomial}
		%      \label{fig:ref_traj_random}
		%    \end{subfigure}%
	%    \hspace*{1mm}
	%    \begin{subfigure}{.23\textwidth}
		%      \centering
		%      \includegraphics[width=1\linewidth]{figures/trajectory_circle.pdf}
		%      \caption{Circular trajectory with radius $r= 10m$ with velocity increasing linearly from $0$ to $v_{\text{max}}$}
		%      \label{fig:ref_traj_circle}
		%    \end{subfigure}
	%    \caption{Reference trajectories used for the experiments.}
	%    \label{fig:ref_trajs}
	%\end{figure}

	\begin{figure}
		\subfloat[Trajectory created by interpolating between randomly generated waypoints with a polynomial]
		{
			\includegraphics[width=0.45\linewidth]{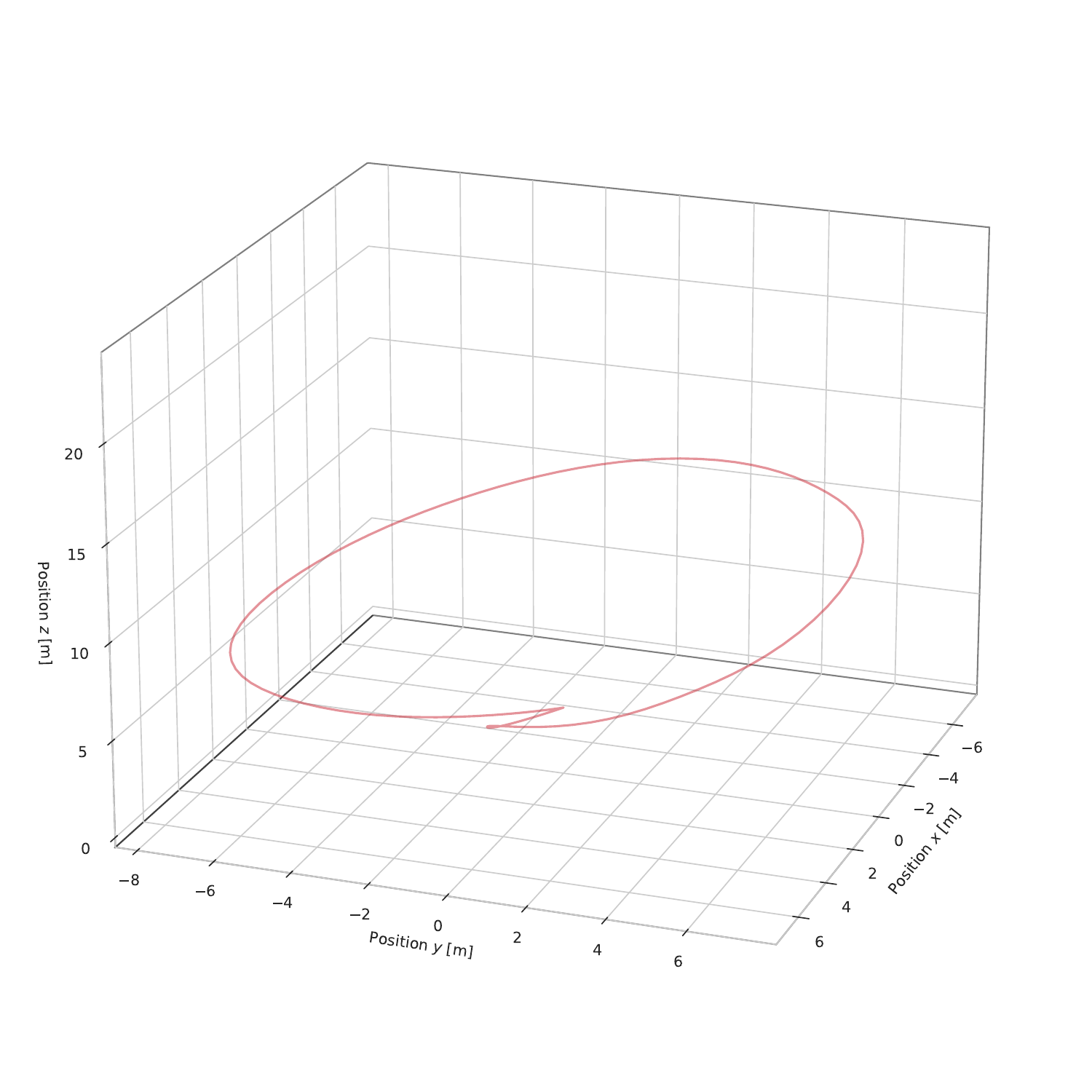}
			\label{fig:ref_traj_random}
		}		\hspace{1mm}
		\subfloat[Circular trajectory with radius $r= 10m$ with velocity increasing linearly from $0$ to $v_{\text{max}}$]
		{
			\centering
			\includegraphics[width=0.45\linewidth]{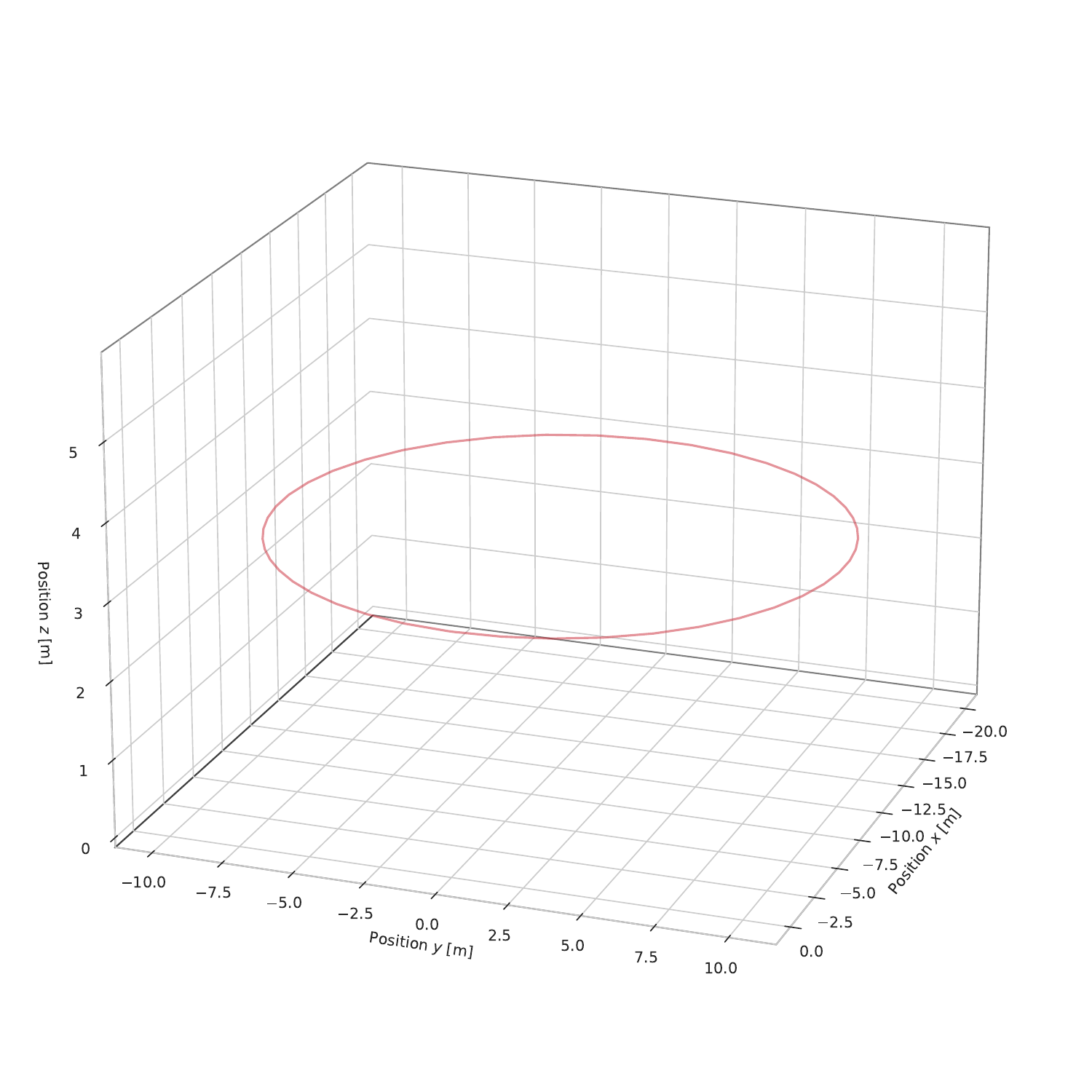}
			\label{fig:ref_traj_circle}
		}
		\caption{Reference trajectories used for the experiments.}
		\label{fig:ref_trajs}
	\end{figure}
	
	\subsection{Nonlinear Model Predictive Control Implementation}
	The MPC algorithm is implemented using the \texttt{Python} interface to \acados/ \citep{verschueren2022acados}, a solver for nonlinear optimal control problems for embedded applications. 
	The system model $\fpred/$ is implemented as a symbolic \casadi/ \citep{andersson2019casadi} function which is used to construct the OCP for the MPC, for which \acados/ generates a solver.

	The MPC is parametrized with a desired horizon look--forward time $t_h$ and the number of prediction steps $n_h$. 
	The length of one prediction step is thus given by $T_h = \frac{t_h}{n_h}$.

	To solve the MPC problem at each time step, we use the \emph{Sequential quadratic programming real-time iteration} (\sqprti/) solver type \citep{verschueren2022acados}. 
	This solver attempts to solve the optimization problem iteratively, but instead of iterating until a desired optimality condition is satisfied, it performs only a single iteration and then returns the current solution.
	This non-optimal solution is then used instead of the optimal quadrotor control.

	\subsection{Experimental Setup}
	The experiments were performed using the \texttt{RotorS} \citep{Furrer2016} package for the Gazebo simulator \citep{koenig2004design} on the circle trajectory shown in \figref{fig:ref_traj_circle} and on a randomly pre-generated trajectories shown in \figref{fig:ref_traj_random}. 
	We set the parameters of the MPC control algorithm as follows, horizon length $t_h = \si{s}$, $n_{h} = 5$ and control interval of $\Delta t = 0.01\si{s}$.  Both the GP and the RGP models used $m=20$ inducing points. 
	
	The GP in the \namegp/ approach was pre-trained by selecting the $m$ inducing points from a dataset generated in a training run on the \emph{random} trajectory using Gaussian mixture modelling (GMM). The hyperparameters of the GP $\eta$ were then optimized using the maximum likelihood estimation method \citep{ljung1998system}.

	The RGP in the \namergp/ approach was not pre-trained before performing the trajectory, and was trained online only. Its $m$ basis vectors were uniformly chosen on $[-v_{\text{max}}, v_{\text{max}}]$ with corresponding initial drag acceleration estimates being $0$.
	Since the RGP does not perform hyperparameter estimation during learning, its hyperparameters  were set at the start of the trajectory as $\eta = \left(1.0,0.1,0.1 \right)$. 
	The need to set hyperparameters beforehand limits the \namergp/ approach, as discussed in \ref{sec:conclusions}.

	\subsection{Results}
	\label{sec:results}
	Having a description of the simulation environment, the simulation results of the experiments performed in Gazebo are discussed further. 
	
	\renewcommand{\arraystretch}{1.0}

	\begin{table}[ht]
		\footnotesize
		\centering
		\captionsetup{width=1.0\linewidth}
		\caption{RMSE position error for the circle trajectory generated using Gazebo with both $\fphys/$ and  $\fpred/$. Both the pretrained GP and the RGP augmented controllers have consistently lower RMSE position error at the cost of increased optimization time.}
		\label{tab:rmsepos}
		\hspace*{-11mm}
		\resizebox{0.9\linewidth}{!}{
			\begin{tabularx}{\linewidth}{l l l l l}
				\cline{1-5}
				& & \multicolumn{3}{c}{\textbf{RMSE pos [mm]}} \\
				\cline{3-5}
				\textbf{Trajectory} & $\bm{v}_{\text{max}}$ [$\text{ms}^{-1}$] & \textbf{Nominal} & \textbf{\namegp/} & \textbf{\namergp/} \\
				\cline{1-5}
				\cline{1-5}
				\multirow{4}*{Random} & 3 & 75.9 & 30.9 {\textbf{(41\%)}} & 40.6 {(53\%)} \\
				& 6 & 110.1 & 52.9 {\textbf{(48\%)}} & 65.1 {(59\%)} \\
				& 9 & 128.5 & 70.3 {\textbf{(55\%)}} & 88.2 {(69\%)} \\
				& 12 & 142.9 & 81.9 {\textbf{(57\%)}} & 99.6 {(69\%)} \\
				\cline{1-5}
				\multirow{4}*{Circle} & 3 & 57.5 & 22.5 {\textbf{(39\%)}} & 23.6 {(41\%)} \\
				& 6 & 102.7 & 50.5 {(49\%)}& 43.5 {\textbf{(42\%)}} \\
				& 9 & 145.1 & 88.1 {(61\%)} & 69.7 {\textbf{(47\%)}} \\
				& 12 & 183.9 & 128.4 {(70\%)}& 98.2 {\textbf{(53\%)}} \\
				\cline{2-5}
				& opt. $\Delta t$ [ms] & 0.60 & 0.66 & 1.21  \\
				\cline{1-5}
			\end{tabularx}
		}
	\end{table}
	
	The results shown in Table \ref{tab:rmsepos} indicate that both the \namegp/ and \namergp/ approaches \emph{significantly} reduce the RMSE position tracking error in comparison to the \emph{nominal}, unaugmented physics-based $\fphys/$ MPC. Performance of MPCs with the augmented models is rather comparable, but the proposed \namergp/ do \emph{not} rely on any pre-training.
	On the other hand, both \namegp/ and \namergp/ increase the optimization time needed to calculate the next MPC output\footnote{Simulations performed on Ubuntu 20.04 with an Intel Core i5-8300H CPU and 8GB of RAM.}. However, this increase is more significant for \namergp/, since it performs additional computations to update the RGP and then to re-parameterize the OCP.

	%Fig. \ref{fig:covariance_v_e} shows the absolute value of covariance between the velocity $v_d$ and the position tracking error $e_d$ for each $d \in \{x,y,z\}$ based on the peak velocity during the simulated trajectory $v_{\text{peak}}$. It can be seen, that the \namergp/ approach reduces this value compared to the \emph{nominal} approach. This is due to the fact that our approach is able to account for the model descrepancy during flight and fit the $\frgp/$ model to compensate for them. Thanks to this compensation, the augmented model $\fpred/$ is makes more accurate predictions of the future state in the MPC control loop, thus minimizing the tracking error compared to the unaugmented model. 
	%
	%\begin{figure}[ht]
	%    \centering
	%    \includegraphics[width=\linewidth]{figures/covariance_data.pdf}
	%    \caption{Absolute values of covariances between the peak tracking velocity $v_{\text{peak}}$ and position tracking error $e_{x,y,z}$ for each of the three dimension $x,y,z$ measured on the circle trajectory captured using the Gazebo simulation.}
	%    \label{fig:covariance_v_e}
	%\end{figure}

\begin{figure}[ht]
		\centering 
		\includegraphics[width=\linewidth]{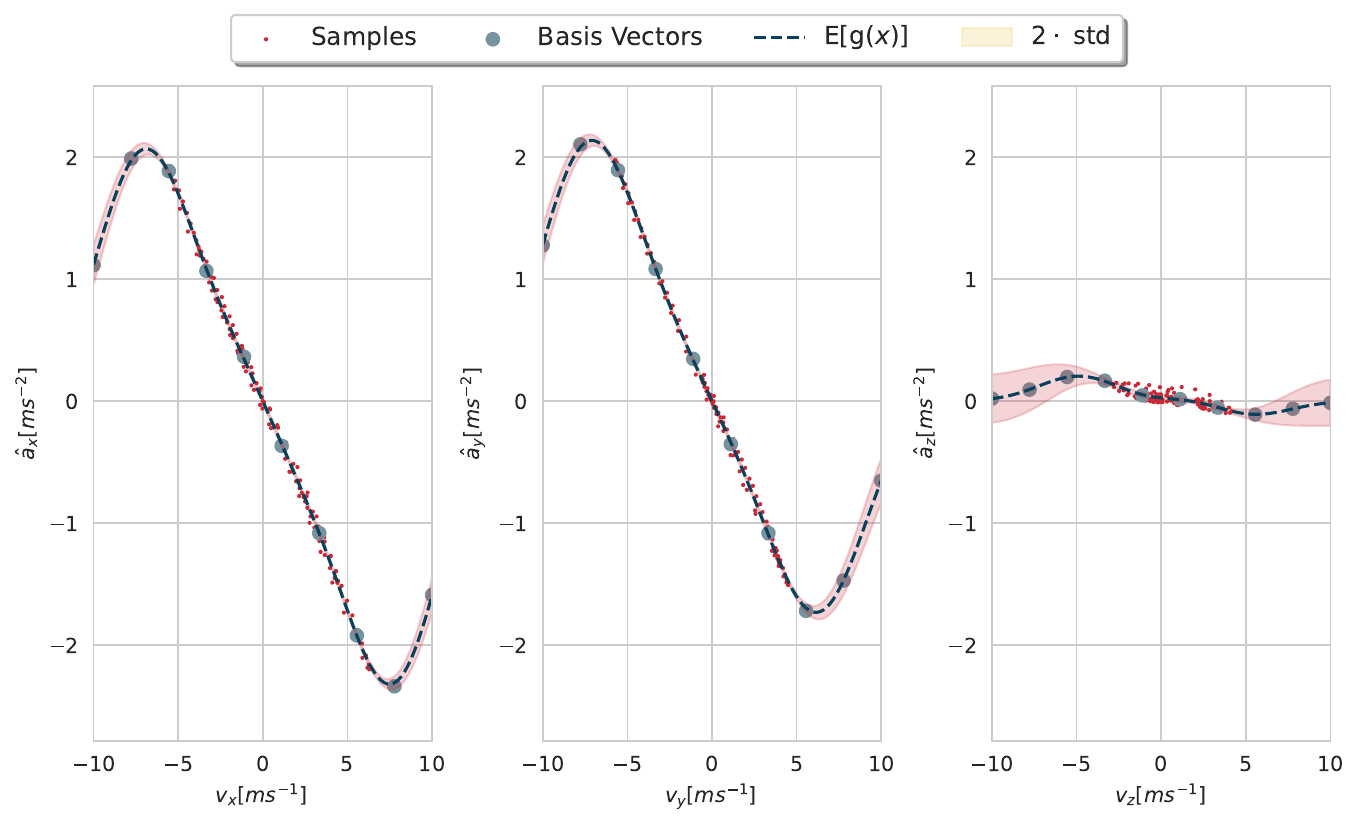}
		\caption{RGP inference space at the end of a trajectory using the Gazebo simulator. Since the Gazebo simulator only models rotor drag \eqref{eq:drag_rotor}, the RGP predicts much lower drag in the $z_B$ direction.} 
		\label{fig:RGP_after}
	\end{figure}

	Fig. \ref{fig:RGP_after} visualises the RGP inference distribution at the   end of a simulation. The RGP is at each time step being incrementaly changed to fit incoming observations and, at the same time, it is being used to make predictions in the MPC control loop.
	
	%Fig. \ref{fig:RGP_before_after} visualises the RGP inference distribution at the start (upper plot) and the end (bottom plot) of a simulation.The RGP is at each time step being incrementaly changed to fit incoming observations and, at the same time, it is being used to make predictions in the MPC control loop.
	%
	%\begin{figure}[ht]
	%    \centering 
	%    \includegraphics[width=\linewidth]{figures/rgp_python_simulation_traj0_v10_a10_gp2_before.pdf}
	%    \par\medskip
	%    \includegraphics[width=\linewidth]{figures/rgp_python_simulation_traj0_v10_a10_gp2_after.pdf}
	%    \caption{RGP inference space at the start (top) and at the end (bottom) of the trajectory using the Gazebo simulator. Since the Gazebo simulator only models rotor drag \eqref{eq:drag_rotor}, the RGP predicts much lower drag in the $z_B$ direction.} 
	%    \label{fig:RGP_before_after}
	%\end{figure}
	%

%%%%%%%%%%%%%%%%%%%%%%%%%%%%%%%%%%%%%%%%%%%%%%%%%%%%%%%%%%%%%%%%%%%%%%%%%%%%%%%%
	\section{CONCLUSIONS AND FUTURE WORK}
	\label{sec:conclusions}
	
	\subsection{Observations, Notes, and Future Work}

	Since we are using estimated disturbance acceleration $\tildea/$ calculated in the $\Bodyframe/$ frame of the quadrotor to fit $\frgp/$, we only have a limited ability to distinguish disturbances that are not time-invariant with respect to $\Bodyframe/$. For example, even for a constant wind in $\Worldframe/$, the disturbance acceleration $\tildea/$ will be time-varying with respect to $\Bodyframe/$ due to the rotation of the quadrotor. The presence of such seemingly time-varying disturbances can induce adaptation that causes increasing $\tildea/$, which might potentially lead to control loop instability.

	While the recursive Gaussian process model provides a uncertainity estimate, we do not utilize this information in a systematic way. 
	The RGP uncertainity is Gaussian, but by propagating it through the nonlinear function in the MPC control loop, this property is lost and theoretical investigation of the resultant probability distribution would need to be performed. Solving this problem would allow one to utilize uncertainty to improve control performance. At current time, the designer has access to the standard deviation of the RGP model and can use it to evalute the goodness of fit. For a more thorough investigation of GP uncertainity propagation through MPC, see \citep{Bradford2020}.

	In the current implementation, we do not perform any hyperparameter optimization in the \namergp/ approach. We found the hyperparameter estimation method \citep{huber2014recursive} augmenting the RGP state by the hyperparameters $\eta$ and estimating them using the Kalman filter to be too unstable to be used safely. Thus, the suitable hyperparameters were selected a priori. On-line adaptation can further increase the ability of the proposed \namergp/ method.
	
	We use the estimated disturbance acceleration \eqref{eq:drag_body_acc} caused by the drag to fit the RGP model without any pre-processing. 
	This can lead to unstable control behaviour when the real quadrotor state differs from the predicted state due to faulty or rare-normal behaviour. 
	A possible solution can be based on  gathering data in a training run and then selects the data points valid for ``nominal'' conditions to fit the GP model in \namegp/ \citep{torrente2021data}. 
	
	\subsection{Conclusions}
	This paper dealt with the predictive control of a quadrotor described by the  data-augmented physics-based model. The data-driven part of the model is realised by a Recursive Gaussian Process regression model that gives an estimate of the air drag force learned online without the need to gather data in a training run. 
	During each control interval, we calculate the residual air drag acceleration as the difference between the commanded and measured velocities divided by the time interval. 
	This residual, together with the current velocity, is used to update the RGP model. 
	The updated RGP model is then used by the MPC to predict the future position of the quadrotor, where the drag force impact is inherently respected. 
	We have shown in realistic simulations that the proposed method is able to quickly adapt to never-before-seen velocities and achieves better position tracking performance than the MPC with the non-augmented (physics-based) model. 
	
	%\citep{Abl:45}

	%The authors gratefully acknowledge the contribution of National Research Organization and reviewers' comments.

	%%%%%%%%%%%%%%%%%%%%%%%%%%%%%%%%%%%%%%%%%%%%%%%%%%%%%%%%%%%%%%%%%%%%%%%%%%%%%%%%
%	

	\small	
	\bibliography{bibliography}
\end{document}